\documentclass[10pt, a4paper]{article}
\usepackage{lrec}
\usepackage{multibib}
\newcites{languageresource}{Language Resources}
\usepackage{graphicx}
\usepackage{tabularx}
\usepackage{soul}
\usepackage{booktabs}
\usepackage{epstopdf}
\usepackage[utf8]{inputenc}

\usepackage{hyperref}
\usepackage{xstring}
\usepackage{multirow}
\usepackage{multicol}
\usepackage{color}
\usepackage{amsmath}
\usepackage{bbm}

\title{HypoNLI: Exploring the Artificial Patterns of Hypothesis-only Bias in Natural Language Inference}

\name{Tianyu Liu\textsuperscript{1}, Xin Zheng\textsuperscript{3}, Baobao Chang\textsuperscript{1,2}, Zhifang Sui\textsuperscript{1,2}}

\address{\textsuperscript{1}MOE Key Lab of Computational Linguistics, School of EECS, Peking University \\
\textsuperscript{2}Peng Cheng Laboratory, Shenzhen, China
\textsuperscript{3}Beijing University of Posts and Telecommunications\\
         tianyu0421@pku.edu.cn, zheng\_xin@bupt.edu.cn, chbb@pku.edu.cn, szf@pku.edu.cn}

\author{Tianyu Liu\textsuperscript{1}, Fuli Luo\textsuperscript{1}, Pengcheng Yang\textsuperscript{1}, Wei Wu\textsuperscript{1}, Baobao Chang\textsuperscript{1,2} \and Zhifang Sui\textsuperscript{1,2} \\
\textsuperscript{1}MOE Key Lab of Computational Linguistics, School of EECS, Peking University \\
\textsuperscript{2}Peng Cheng Laboratory, Shenzhen, China \\
  {\tt \{tianyu0421, luofuli, yang\_pc, wu.wei, chbb, szf\}@pku.edu.cn}}

\abstract{
Many recent studies have shown that for models trained on datasets for natural language inference (NLI), it is possible to make correct predictions by merely looking at the hypothesis while completely ignoring the premise. 
In this work, we manage to derive adversarial examples in terms of  the hypothesis-only bias and explore eligible ways to mitigate such bias. 
Specifically, we extract various phrases from the hypotheses (artificial patterns) in the training sets, and show that they have been strong indicators to the specific labels. We then figure out `hard' and `easy' instances from the original test sets whose labels are opposite to or consistent with those indications.
We also set up baselines including both pretrained models (BERT, RoBERTa, XLNet) and competitive non-pretrained models (InferSent, DAM, ESIM).
Apart from the benchmark and baselines, we also investigate two debiasing approaches which exploit the artificial pattern modeling to mitigate such hypothesis-only bias: down-sampling and adversarial training.
We believe those methods can be treated as competitive baselines in NLI debiasing tasks.
\newline \Keywords{Natural Language Inference, Hypothesis-only Bias, Artificial Patterns} }

\begin{document}

\maketitleabstract

\section{Introduction}
Natural language inference (NLI) (also known as recognizing textual entailment) is a widely studied task which aims to infer the relationship 
(e.g., \emph{entailment}, \emph{contradiction}, \emph{neutral})
between two fragments of text, known as \emph{premise} and \emph{hypothesis} \cite{dagan2006pascal,DBLP:series/synthesis/2013Dagan}. NLI models are usually required to determine whether a hypothesis is true (\emph{entailment}) or false (\emph{contradiction}) given the premise, or whether the truth value can not be inferred (\emph{neutral}).
A proper NLI decision should apparently rely on both the premise and the hypothesis. However, some recent studies \cite{gururangan2018annotation,poliak2018hypothesis,tsuchiya2018performance} have shown that it is possible for a trained model to identify the true label by only looking at the hypothesis without observing the premise. 
The phenomenon is referred to as annotation artifacts \cite{gururangan2018annotation}, statistical irregularities \cite{poliak2018hypothesis} or partial-input heuristics \cite{feng2019misleading}.
In this paper we use the term \emph{hypothesis-only bias} \cite{poliak2018hypothesis} to refer to this phenomenon.

Such hypothesis-only bias originates from the human annotation process of data collection. In the data collection process of many large-scale NLI datasets such as SNLI \citelanguageresource{snli:emnlp2015} and MultiNLI \citelanguageresource{mnli:N18-1101}, human annotators are required to write new sentences (hypotheses) based on the given premise and a specified label among \emph{entailment}, \emph{contradiction} and \emph{neutral}. Some of the human-elicited hypotheses contain patterns 
that spuriously correlate to some specific labels.
For example, 85.2\% of the hypothesis sentences which contain the phrase \textit{video games} were labeled as \emph{contradiction}. The appearance of \textit{video games} in hypothese can be seen as a stronger artificial indicator to the label \emph{contradiction}.

To get a deeper understanding of the specific bias captured by NLI models in the training procedure, we try to extract explicit surface patterns from the training sets of SNLI and MultiNLI, and show that the model can easily get decent classification accuracy by merely looking at these patterns.
%The release of large-scale NLI datasets, such as SNLI \cite{snli:emnlp2015} and MultiNLI \cite{mnli:N18-1101}, makes complex deep learning methods applicable on this task.
%In the data collection process, human annotators are required to generate new sentences (hypotheses) based on the given premise and the gold label. 
%The human-elicited hypotheses might give away the gold label, which is referred to as annotation artifacts \cite{gururangan2018annotation} or statistical irregularities \cite{poliak2018hypothesis}. In this paper, we use the term 'hypothesis-only bias' to refer to .
After that, we derive hard (adversarial) and easy subsets from the original test sets. They are derived based on the indication of the artificial patterns in the hypotheses. The gold labels of easy subsets are consistent with such indication while those of hard subsets are opposite to such indication.
The model performance gap on easy and hard subsets shows to what extend a model can mitigate the hypothesis-only bias.

\begin{table*}[t]
\small
\begin{center}
\setlength{\tabcolsep}{0.9mm}{
\begin{tabular}{cllllll|llllll}
\toprule
 & \multicolumn{6}{c}{Multi-word Patterns} & \multicolumn{6}{|c}{Unigram Patterns}\\
 & \multicolumn{2}{c}{Entailment} & \multicolumn{2}{c}{Neutral}  & \multicolumn{2}{c}{Contradiction} & \multicolumn{2}{|c}{Entailment} & \multicolumn{2}{c}{Neutral}  & \multicolumn{2}{c}{Contradiction} \\
 \midrule
\multirow{3}{*}{SNLI} & in this picture & 96.4 & tall human & 99.7 & Nobody \# \# . & 99.8  & outdoors & 78.8 & vacation & 91.0 & Nobody & 99.7 \\
 & A human	& 96.4 & A sad & 95.6 & dog \# sleeping & 97.5  & sport & 75.1 & winning & 89.9 & No & 95.8 \\
 & \underline{A \# \# outdoors .}  & 95.9 & A \# human & 94.1 & There \# no & 96.2 & instrument & 74.4 & favorite & 88.7 & cats & 93.4\\
 & A \# \# outside . & 89.8 & the first & 88.6 & in \# bed & 94.2 & animal & 68.5 & date & 87.4 & naked & 88.7 \\
 & is near \# \# . & 87.6 & on \# way & 87.0 & at home & 93.5 & moving & 67.8 & brothers & 85.6  & tv & 88.4\\
\midrule
\multirow{3}{*}{MultiNLI}& It \# possible & 71.7 & , said the & 93.6 & There are no & 92.4 & Several & 54.7 & addition & 69.6  & None & 85.4\\
 & There \# a \# \# the  & 70.8 & They wanted to & 81.4 & does not \# any & 91.9 & Yes & 54.4 & also & 68.6  & refused & 80.5\\
 & There is an & 68.8 & the most popular & 78.7 & no \# on & 91.5 & various & 53.7 & locals & 65.7  & never & 79.0\\
 & are two & 67.0 & addition to & 78.4 & are \# any & 90.1 & $\cdots$ & 53.1 & battle & 63.3  & perfectly & 77.3\\
 & There \# some & 65.9 & because he was & 77.8 & are never & 89.9 & According & 53.1 & dangerous & 63.2  & Nobody & 77.1\\
\bottomrule
\end{tabular}}
\end{center}
\caption{Top 3 artificial patterns sorted by the pattern-label conditional probability $\mathrm{p}(l|b)$ (Sec \ref{extraction}). The listed patterns appear at least in 500/200 instances in SNLI/MultiNLI training sets, notably the numbers 500/200 here are chosen only for better visualization. `\#' is the placeholder for an arbitrary token.
The \underline{underlined} artificial pattern serves as an example in Sec \ref{extraction}.}\label{showpatterns}
%For multi-word patterns, we filter the patterns which might have overlap, e.g. we do not show the pattern `said the \# .' as it may have overlaps with `, said the'. 
\end{table*}

After analyzing some competitive NLI models, including both non-pretrained models like Infersent \cite{conneau2017supervised}, DAM \cite{parikh2016decomposable} and ESIM \cite{chen2016enhanced} and popular pretrained models like BERT \cite{devlin2018bert}, XLNet \cite{yang2019xlnet} and RoBERTa \cite{liu2019roberta}, we find that the hypothesis-only bias makes NLI models vulnerable to the adversarial (hard) instances which are against such bias (accuracy $<$ 60\% on InferSent), while these models get much higher accuracy (accuracy $>$ 95\% on InferSent) on the easy instances.
This is an evidence to show that the NLI models might be over-estimated as they benefit a lot from the hints of artificial patterns. 
%Worse still, we observe that such bias even hurts the model accuracy on the unbiased subsets. So we consider it meaningful to investigate how we can mitigate such bias.

A straightforward way is to eliminate these human artifacts in the human annotation process, such as encouraging human annotators to use more diverse expressions or do dataset adversarial filtering \cite{zellers2018swag} and multi-round annotation \cite{nie2019adversarial}. However in this way, the annotation process would inevitably become more time-consuming and expensive.
%as we need to warn the annotators not to use the specific ways of expression by dynamically deriving the potential artificial patterns. 

To this end, this paper explores two ways based on the derived artificial patterns to alleviate the hypothesis-only bias \emph{in the training process}: down-sampling and adversarial debiasing.
We hope they would serve as competitive baselines for other NLI debiasing methods.
Down-sampling aims at reducing the hypothesis-only bias in NLI training sets by removing those instances in which the correct labels may easily be revealed by artificial patterns.
Furthermore, we exploring adversarial debiasing methods \cite{belinkov2019don,belinkov2018mitigating} for the sentence vector-based models in NLI \cite{yang2016hierarchical,lin2017structured,wu2018phrase,luo2018leveraging}. The experiments show that the guidance from the derived artificial patterns can be helpful to the success of sentence-level NLI debiasing.

\section{Datasets}
%We hypothesize that the hypothesis-only bias is mainly caused by fitting certain surface patterns. 
In this section, we identify the artificial patterns from the hypothesis sentences which highly correlate to specific labels in the training sets and 
then derive hard, easy subsets from the original test sets based on them.  
%verify that current models are relying heavily on the bias patterns to make correct predictions (Sec \ref{causeofbias}).
%To get a deeper understanding on the hypothesis-only bias, we try to extract multi-word patterns and investigate whether these patterns are the main underlying factors for the hypothesis-only bias.  
\subsection{Artificial Pattern Collection}\label{extraction}
`Pattern' in this work refers to (maybe nonconsecutive) word segments in the hypothesis sentences. We try to identify the `artificial patterns' which spuriously correlate to a specific label due to certain human artifacts. 

We use $\mathrm{H}(M,t,\lambda)$ to represent a set of artificial patterns. $M$ and $t$ denotes the max length of the pattern and the max distance between two consecutive words in a pattern, respectively. For a artificial pattern $b\in \mathrm{H}(M,t,\lambda)$, there exists a specific label $l$ for $b$ that the conditional probability $\mathrm{p}(l|b)= count(b,l) / count(b)>\lambda$. For example, for the underlined pattern `A \# \# outdoors .' in Table \ref{showpatterns}, the length of this pattern is 3, and the distance between the consecutive words `A' and `outdoors' is 2. Its conditional probability with the label \emph{entailment} is 95.9\%.
Notably, all the recognized artificial patterns in our paper appear in at least 50 instances of the training sets to avoid misrecognition\footnote{Suppose a pattern only appears once in a training instance, its $\mathrm{p}(l|b)$ always equals 1 for the label in that instance.}.

In the rest of paper, unless otherwise specified, we set $M=3,t=3$ \footnote{We also tried larger $M$ and $t$, e.g. 4 or 5, but did not observe considerable changes of the artificial patterns, e.g. 95.4\% patterns in H(5,5,0.5) are covered by H(3,3,0.5).\label{fn:mandt}}. By doing so, we only tune the hyper-parameter $\lambda$ in $\mathrm{H}$(3,3,$\lambda$) to decide using a rather strict (smaller $\lambda$) or mild (bigger $\lambda$) strategy while deriving the artificial patterns.

\subsection{Analysis of Hypothesis-only Bias}\label{causeofbias}
Previous work \cite{gururangan2018annotation,poliak2018hypothesis} trained a sentence-based hypothesis-only classifier which achieves decent accuracy.
Different from them, we show that in Table \ref{classifier} the classifier which merely uses the artificial patterns as features achieves comparable performance with the fasttext \cite{Joulin2016FastText} classifier.
%Our intention of capturing the artificial patterns is to figure out specific ways of expression of the annotators that causes the hypothesis-only bias, thus different from previous work which only concern unigram words \cite{gururangan2018annotation,poliak2018hypothesis}, we use multi-word patterns (including unigram patterns) in our paper.
Table \ref{classifier} shows the classifier based on multi-word patterns with the default $M$ and $t$ (see Footnote \ref{fn:mandt}, $\mathrm{H}$(3,3,0.5)) achieves much higher accuracy than that based on only unigram patterns ($\mathrm{H}$(1,1,0.5)).

\begin{table}[t]
\small
\begin{center}
\begin{tabular}{cccc}
\toprule
\bf \multirow{2}{*}{Model}  & \bf \multirow{2}{*}{SNLI}  
&  \multicolumn{2}{c}{\bf MultiNLI}  \\
 & & Matched & Mismatched \\ \midrule
majority class & 34.3 & 35.4 & 35.2 \\
\texttt{fasttext} & 67.2 & 53.7 & 52.5 \\
Unigram & 60.2 & 49.8 & 49.6 \\
Pattern & 64.4  & 52.9 & 52.9 \\
\bottomrule
\end{tabular}
\end{center}
\caption{The accuracies of the hypothesis-only classifiers on SNLI test and MultiNLI dev sets. 
%The first two classifiers are tested by \citet{gururangan2018annotation}. 
We train a MLP classifier with unigrams (Unigram) or multi-words patterns (Pattern) as features. Details in Sec \ref{causeofbias}.}\label{classifier}
\end{table}

We also compare the test accuracies on the easy and hard sets (Sec \ref{hardandeasy}) of the baseline models (I-9,D-9,E-9) in Table \ref{tb:train_on_snli} and \ref{tb:train_on_mnli}. Empirically we find that the NLI models achieve very high accuracy on the easy sets while performing poorly on the hard sets. We also observe the same tendency in the models trained on the randomly downsampled training sets (e.g. I-1, I-3, I-5, I-7, etc.). It shows that NLI models fit the artificial patterns in the training set very well, which makes them fragile to the adversarial examples (hard set) which are against these patterns. Thus we assume the artificial patterns contributes to the hypothesis-only bias.

\subsection{Hard and Easy Subsets}\label{hardandeasy}

Some instances contain artificial patterns that are strong indicators to the specific labels. We treat the instances in the test sets which are consistent with such indication as `easy' ones and those instances which are against such indication as `hard' ones.

For easy subsets, the labels of \emph{all} the artificial patterns in the specific hypotheses must be consistent with the gold labels. We show an easy instance below: the artificial patterns `The dogs are \# on' and `bed .' (bed is the last word of the sentence) are strong indicators to the correct classification.

\begin{table}[]
    \centering
    \begin{tabular}{l}
\toprule
\textbf{Premise}:  Two cats playing on the bed together . \\
\textbf{Hypothesis}:  \emph{The dogs are} playing \emph{on} the \texttt{bed .}  \\
\textbf{Gold Label}:  \textsc{Contradiction} \\
\textbf{Artificial patterns}:
(\texttt{bed .}, \textsc{Contradiction}, 83.2\% \\
);
(\emph{The dogs are \# on}, \textsc{Contradiction}, 82.9\%) \\
\bottomrule
    \end{tabular}
    \\(a) An easy instance
\begin{tabular}{l}
\toprule
\noindent \textbf{Premise}:  A bare-chested man fitting his head and arm\\
 into a toilet seat ring while spectators watch in a city.\\
\textbf{Hypothesis}:  A gentleman with \texttt{no} chest hair , \\
wrangles \emph{his way} through a toilet seat . \\
\textbf{Gold Label}:  \textsc{Entailment} \\
\textbf{Artificial patterns}: (\texttt{no}, \textsc{Contradiction}, 82.7\%)\\
(\emph{his way}, \textsc{Neutral}, 82.4\%) \\
\bottomrule
\end{tabular}
\\(b) A hard instance
\caption{Examples for easy and hard instances. The indications of artificial patterns are consistent with the gold label in the easy case while they are against the gold label in the hard case. The triple $(\mathbf{P}, l, p)$ show the related label indications $l$ for specific artificial patterns $\mathbf{P}$ and their conditional probability $p$.}
\end{table}

For the hard subsets, on the other side, the indications of the artificial patterns should be \emph{all} different from gold labels. We also show a hard instance below: in this situation, the artificial patterns `no'\footnote{`no' is different from `No' shown in Table \ref{showpatterns} as the latter indicates the word appears in the beginning of the sentence. } and `his way' may misguide the NLI models to the wrong answers.

Notably we do not put instances with conflicting indications (e.g. an instance with 2 artificial patterns, one of which has the same label with the gold label while the other does not) into easy or hard subsets to build more challenging adversarial examples.

The sizes of hard and easy sets actually depend on how we harvest artificial pattern, i.e. $\lambda$ in $\mathrm{H}(M=3,t=3,\lambda)$ (Sec \ref{extraction} and Footnote \ref{fn:mandt}).
For the sake of simplicity, we utilize $\lambda=0.8$ and $\lambda=0.7$ \footnote{MultiNLI's pattern-label conditional possibilities are generally smaller than those of SNLI as shown in Table \ref{showpatterns}. So we use smaller $\lambda$ to ensure the size of derived subsets. } as the thresholds to derive easy and hard subsets for SNLI and MultiNLI respectively in the following experiments, as adopting a relatively bigger $\lambda$ can choose the instances which largely accord with the artificial patterns and are thus eligible to serve as adversarial examples.

The sizes of easy and hard sets in SNLI test set, MultiNLI-matched dev set and MultiNLI-mismatched dev set are 327/1760; 410/1032; 371/1085 respectively. \footnote{The datasets used in this paper can be found in https://tyliupku.github.io/publications/}
The performance of an ideally unbiased NLI model on the easy and hard sets should be close to each other. Besides we should not see huge gap between the model accuracy on the easy and hard subsets.

\subsection{Baselines}\label{sec:baseline}
We set up both pretrained and non-pretrained model baselines for the proposed adversarial datasets. We rerun their public available codebase with the default hyper-parameter and optimizer settings, including InferSent\footnote{https://github.com/facebookresearch/InferSent}, DAM\footnote{https://github.com/harvardnlp/decomp-attn}, ESIM\footnote{https://github.com/coetaur0/ESIM}, BERT (uncased base), XLNet (cased base) and RoBERTa (base)\footnote{https://github.com/huggingface/transformers}. 
For BERT, XLNet and RoBERTa, we concatenate the premise sentence and hypothesis sentence with [SEP] token as the input. For output classifier, we use a linear mapping to transform the vector at the position of [CLS] token at the last layer of these pretrained models to a normalized 3-element vector (using softmax) which represents the scores for each label. 
We report the test accuracies on easy, hard subsets and the UW+CMU hard subsets \cite{gururangan2018annotation} which are derived from a hypothesis-only classifier. From Table \ref{tab:big_baseline}, we can tell that the proposed hard sets are more challenging than UW+CMU hard subsets.

\section{Exploring Debiasing Methods}
\begin{table}[]
    \centering
    \begin{tabular}{lcccc}
\toprule
          & Full & Easy & Hard & UW+CMU  \\ \midrule
InferSent & 84.5 & 97.2 & 58.9 &  69.3\\
DAM & 85.8 & 97.8 & 62.1 & 72.0\\
ESIM & 87.6 & 97.7 & 68.2 & 75.2\\ \midrule
BERT & 90.5 & 98.2 & 71.2 & 80.3 \\
XLNet & 90.9 & 98.0 & 73.6 & 80.7 \\
RoBERTa & 91.7 & 98.9 & 75.8 & 82.7 \\ \bottomrule
    \end{tabular}
    \\(a) Models trained on SNLI
    \begin{tabular}{lcccc}
\toprule
          & Full & Easy & Hard \\ \midrule
InferSent & 70.4 & 92.7 & 54.4\\
DAM & 70.5 & 92.0 & 55.1\\
ESIM & 76.7 & 93.9 & 65.6\\ \midrule
BERT & 83.4 & 95.2 & 75.0\\
XLNet & 86.5 & 96.3 & 78.2 \\
RoBERTa & 87.2 & 96.5 & 81.4  \\ \bottomrule
    \end{tabular}
    \\(b) Models trained on MultiNLI
\caption{Model baselines on the proposed hard and easy test sets. For MultiNLI, we trained the models using matched dev sets as the valid set and reported the test accuracies on mismatched dev sets. 'UW+CMU' refers to the adversarial set (Sec \ref{sec:baseline}) detected by a neural based hypothesis-only classifier.}
    \label{tab:big_baseline}
\end{table}

\subsection{Down-sampling Baselines}

% trained on SNLI
\begin{table}[!t]
\small
\begin{center}
\begin{tabular}{ccccccc}
\hline 
$\lambda$ & No. & Mode & Full & Easy & Hard & $\Delta^{Hard}_{Easy}(\downarrow)$ \\ \hline
\multirow{2}{*}{0.5} & I-1 & Rand & \textbf{76.4} & \textbf{93.8} & 48.7 & 45.1\\
& I-2 & Debias & 66.9 & 64.6 & \textbf{56.0} & 8.6  \\ \hline
\multirow{2}{*}{0.6} & I-3 & Rand & \textbf{81.1} & \textbf{96.1} & 54.1 & 42.0 \\
& I-4 & Debias & 76.9 & 79.8 & \textbf{58.0} & 21.8  \\ \hline
\multirow{2}{*}{0.7} & I-5 & Rand & \textbf{82.8} & \textbf{96.9} & 56.0 & 40.9 \\
& I-6 & Debias & 80.9 & 86.4 & \textbf{\underline{61.5}} & 24.9  \\ \hline
\multirow{2}{*}{0.8} & I-7 & Rand & \textbf{83.5} & \textbf{96.9} & 56.6 & 40.3 \\
& I-8 & Debias & 82.9 & 90.4 & \textbf{60.0} & 30.4  \\ \hline
1.0 & I-9 &  All & \underline{84.5} & \underline{97.2} & 58.9 & 38.3  \\
\hline
\end{tabular}
\\(a) InferSent trained on SNLI

\begin{tabular}{ccccccc}
\hline 
$\lambda$ & No. & Mode & Full & Easy & Hard & $\Delta^{Hard}_{Easy}(\downarrow)$ \\ \hline
\multirow{2}{*}{0.5} & D-1 & Rand & \textbf{74.1} & \textbf{92.3} & 46.7 & 45.6\\
& D-2 & Debias & 67.5 & 67.8 & \textbf{53.3} & 14.5\\ \hline
\multirow{2}{*}{0.6} & D-3 & Rand & \textbf{82.4} & \textbf{96.3} & 56.7 & 39.6 \\
& D-4 & Debias & 79.3 & 84.4 & \textbf{62.1} & 21.3 \\ \hline
\multirow{2}{*}{0.7} & D-5 & Rand & \textbf{84.4} & \textbf{97.3} & 59.4 & 37.9 \\
& D-6 & Debias & 83.0 & 89.6 & \underline{\textbf{63.1}} & 26.5  \\ \hline
\multirow{2}{*}{0.8} & D-7 & Rand & \textbf{85.3} & \underline{\textbf{97.8}} & 60.1 & 37.7\\
& D-8 & Debias & 84.6 & 93.5 & \textbf{62.6} & 30.9  \\ \hline
1.0 & D-9 &  All & \underline{85.8} & \underline{97.8} & 62.1 & 35.7  \\
\hline
\end{tabular}
\\(b) DAM trained on SNLI

\begin{tabular}{ccccccc}
\hline 
$\lambda$ & No. & Mode & Full & Easy & Hard & $\Delta^{Hard}_{Easy}(\downarrow)$ \\ \hline
\multirow{2}{*}{0.5} & E-1 & Rand & \textbf{76.8} & \textbf{94.6} & 48.9 & 45.7\\
& E-2 & Debias & 65.3 & 62.5 & \textbf{53.8} & 7.7\\ \hline
\multirow{2}{*}{0.6} & E-3 & Rand & \textbf{83.6} & \textbf{96.6} & 62.2 & 34.4 \\
& E-4 & Debias & 78.6 & 79.4 & \textbf{63.6} & 15.8  \\ \hline
\multirow{2}{*}{0.7} & E-5 & Rand & \textbf{85.9} & \textbf{97.2} & 64.2 & 35.0 \\
& E-6 & Debias & 83.8 & 88.2 & \underline{\textbf{68.8}} & 19.4  \\ \hline
\multirow{2}{*}{0.8} & E-7 & Rand & \textbf{86.9} & \textbf{97.3} & 67.9 & 29.4 \\
& E-8 & Debias & 86.2 & 92.1 & \textbf{70.9} & 21.3  \\ \hline
1.0 & E-9 &  All & \underline{87.6} & \underline{97.6} & 68.2 & 29.4  \\
\hline
\end{tabular}
\\(c) ESIM trained on SNLI
\end{center}
%}
\caption{Model performance on the SNLI test set. We report the average scores of multiple independent runs. 
$\Delta^{Hard}_{Easy}$ represents the gap between hard and easy test sets (lower is better).
$\lambda$ is the debiasing threshold. 
We use 2 `modes' to down-sample the training sets, namely biased instances removing (`Debias') and randomly downsampling  (`Rand'), the latter has the same training size and label distribution with with `Debias' mode for a fair comparison. 
The downsampled training sizes are 4.0\%, 19.8\%, 43.8\%, 67.4\% and 100\% of the whole training size (549867) for $\lambda\in\{0.5,0.6,0.7,0.8,1.0\}$ respectively. Note that when $\lambda$=1.0, we use the whole training set without any downsampling. }\label{tb:train_on_snli}
\end{table}

\begin{table}[!t]
\small
\begin{center}
\begin{tabular}{ccccccc}
\hline 
$\lambda$ & No. & Mode & Full & Easy & Hard & $\Delta^{Hard}_{Easy}(\downarrow)$ \\ \hline
\multirow{2}{*}{0.5} & I-1 & Rand & \textbf{67.5} & \textbf{92.6} & 50.9 & 41.7 \\
& I-2 & Debias & 64.2 & 76.0 & \textbf{56.1} & 19.9  \\ \hline
\multirow{2}{*}{0.6} & I-3 & Rand & \textbf{69.0} & \textbf{92.7} & 53.4 & 39.3 \\
& I-4 & Debias & 67.5 & 80.9 & \underline{\textbf{59.7}} & 21.2  \\ \hline
\multirow{2}{*}{0.7} & I-5 & Rand & \textbf{69.1} & \underline{\textbf{93.0}} & 52.2 & 40.8 \\
& I-6 & Debias & 68.3 & 84.6 & \textbf{57.1} & 27.5  \\ \hline
\multirow{2}{*}{0.8} & I-7 & Rand & 69.2 & \textbf{92.5} & 52.5 & 40.0 \\
& I-8 & Debias & \textbf{69.6} & 91.4 & \textbf{53.6} & 37.8  \\ \hline
1.0 & I-9 &  All & \underline{70.4} & 92.7 & 54.4 & 38.3  \\
\hline
\end{tabular}
\\(a) InferSent trained on MultiNLI

\begin{tabular}{ccccccc}
\hline 
$\lambda$ & No. & Mode & Full & Easy & Hard & $\Delta^{Hard}_{Easy}(\downarrow)$ \\ \hline
\multirow{2}{*}{0.5} & D-1 & Rand & \textbf{64.4} & \textbf{91.9} & 46.2 & 45.7\\
& D-2 & Debias & 61.9 & 77.4 & \textbf{52.0} & 22.4\\ \hline
\multirow{2}{*}{0.6} & D-3 & Rand & \textbf{68.3} & \textbf{91.8} & 52.5 & 39.3 \\
& D-4 & Debias & 65.8 & 79.0 & \underline{\textbf{58.0}} & 21.0  \\ \hline
\multirow{2}{*}{0.7} & D-5 & Rand & \textbf{69.6} & \underline{\textbf{92.6}} & 52.2 & 40.4 \\
& D-6 & Debias & 68.0 & 83.9 & \textbf{57.5} & 25.4  \\ \hline
\multirow{2}{*}{0.8} & D-7 & Rand & \textbf{70.0} & \textbf{92.4} & 53.8 & 38.6 \\
& D-8 & Debias & 69.6 & 91.6 & \textbf{54.9} & 26.7  \\ \hline
1.0 & D-9 &  All & \underline{70.5} & 92.0 & 55.1 & 36.9  \\
\hline
\end{tabular}
(b) DAM trained on MultiNLI

\begin{tabular}{ccccccc}
\hline 
$\lambda$ & No. & Mode & Full & Easy & Hard & $\Delta^{Hard}_{Easy}(\downarrow)$ \\ \hline
\multirow{2}{*}{0.5} & E-1 & Rand & \textbf{70.4} & \textbf{92.8} & 59.0 & 33.8 \\
& E-2 & Debias & 67.0 & 75.2 & \textbf{63.6} & 11.6  \\ \hline
\multirow{2}{*}{0.6} & E-3 & Rand & \textbf{73.9} & \textbf{93.8} & 61.6 & 22.2 \\
& E-4 & Debias & 73.2 & 84.8 & \underline{\textbf{66.4}} & 18.4  \\ \hline
\multirow{2}{*}{0.7} & E-5 & Rand & 74.6 & \textbf{92.8} & 64.5 & 28.3 \\
& E-6 & Debias & \textbf{74.9} & 89.1 & \textbf{65.9} & 23.2  \\ \hline
\multirow{2}{*}{0.8} & E-7 & Rand & 75.7 & \underline{\textbf{94.0}} & 64.4 & 29.6 \\
& E-8 & Debias & \textbf{75.8} & 93.4 & \textbf{65.0} & 28.4  \\ \hline
1.0 & E-9 &  All & \underline{76.7} & 93.9 & 65.6 & 28.3  \\
\hline
\end{tabular}
\\(c) ESIM trained on MultiNLI
\end{center}
%}
\caption{Models performance on MultiNLI mismatched dev set. 
We tune the models on MultiNLI matched dev set. 
The training sizes are 24.4\%, 53.1\%, 68.0\%, 81.2\% and 100\% of the whole training size (392702) for $\lambda\in\{0.5,0.6,0.7,0.8,1.0\}$ respectively. Note that we do not report the scores on MultiNLI test sets as they are unable to access. The \textbf{bold} numbers mark higher scores between `Rand' and `Debias' mode for each $\lambda$. The \underline{underlined} numbers highlight the highest scores in each column.}\label{tb:train_on_mnli}
\end{table}

Sec \ref{causeofbias} verifies that the artificial patterns lead to correct hypothesis-only classification, which motivates us to remove such patterns in the training sets by down-sampling.
Specifically we down-sample the training sets of SNLI and MultiNLI and retrain 3 prevailing NLI models: InferSent, DAM and ESIM. 

\subsubsection{Downsampling Details}
We down-sampled the training sets by removing the biased instances (`Debias' mode) that contain the artificial patterns.

\noindent \textbf{Choosing down-sampling threshold $\lambda$}: 
The threshold $\lambda$ is exactly the same $\lambda$ defined in Sec \ref{extraction}. We consider a training instance as a biased one even if it contains only one artificial pattern. 
%in $\mathrm{H}$(3,3,$\lambda$) (Sec \ref{extraction}, Footnote \ref{fn:mandt}), with different debiasing thresholds ($\lambda \in \{0.5,0.6,0.7,0.8\}$).
When adopting smaller $\lambda$, we harvest more artificial patterns as described in Sec \ref{extraction}. Accordingly more training instances would be treated as biased ones and then filtered. 
In a word, smaller $\lambda$ represents more strict down-sampling strategy in terms of filtering the artificial patterns. 
%use more aggressive way to filter more biased training instances according to the derive bias patterns.
$\lambda=0.5$ serves as the lower bound because the highest pattern-label conditional probability ($\mathrm{p}(l|b)$ in Sec \ref{extraction}) for premises, which aren't observed the same bias as hypotheses, is less than 0.5 in both SNLI and MultiNLI training set..

\noindent \textbf{Ruling out the effects of training size}: The model performance might be highly correlated with the size of training set. To rule out the effects of training size as much as possible, we set up randomly down-sampled training sets (`rand' mode) with the same size as the corresponding `debias' mode under different $\lambda$ for a fair comparison.

\noindent \textbf{Keeping the label distribution balanced}: After removing the biased instances in the training set by different $\lambda$ (`debias' mode), suppose we get $n_1, n_2, n_3$ ($n_1 \ge n_2  \ge n_3$) instances for the 3 pre-defined labels of NLI in the down-sampled training set. Then we down-sample the subsets with $n_1, n_2$ instances to $n_3$ instances and get a dataset with $3n_3$ instances. For the corresponding `rand' mode, we sample $n_3$ instances for each pre-defined label from training set.

\noindent \textbf{Convincing scores of multiple runs}: To relieve the randomness of randomly down-sampling and model initialization, for the `rand' mode in Table \ref{tb:train_on_snli} and \ref{tb:train_on_mnli}, we firstly randomly down-sample the training set (with the label distribution balanced) according to different $\lambda$ for 5 times and get 5 randomly down-sampled training sets for each $\lambda$. Then for each down-sampled training set, we run 3 independent experiments with random model initialization under the same experimental settings. So each score in the `rand' mode of Table \ref{tb:train_on_snli} and \ref{tb:train_on_mnli} comes from 15 independent runs. The scores in the `debias' mode of Table \ref{tb:train_on_snli} and \ref{tb:train_on_mnli} come from 5 independent runs with random model initialization.

\subsubsection{Discussions}\label{implement}
From table \ref{tb:train_on_snli} and \ref{tb:train_on_mnli}, we observe that:
\noindent 1) The NLI models fit the bias patterns in the hypotheses very well even in the small-scale randomly down-sampled training sets (I-1, D-1 and E1) which only accounts for 4.0\% of the original training set (SNLI), as the performance gaps between easy and hard subsets in these settings are still huge ($>$40\% for SNLI in Table \ref{tb:train_on_snli}).

\noindent 2) Under the same $\lambda$, the proposed `debias' down-sampling not only outperforms its `rand' counterpart in terms of hard subsets, but also greatly reduce the performance gap on easy and hard sets.

\noindent 3) The gains on hard sets on MultiNLI are smaller than those on SNLI as MultiNLI is less biased regarding the pattern-label conditional probability (Table \ref{showpatterns}). Down-sampling achieves larger gains on more biased datasets. In SNLI, the `debias' down-sampling even outperforms the baseline models (I-8 vs I-9, D-8 vs D-9, E-8 vs E-9), which is really impressive as the training size of I-8, I-8 and E-8 is only 67.5\% of the baseline models.

\cite{gururangan2018annotation} expressed concerns upon down-sampling (DS) methods: 1) Will removing the artificial patterns cause new artifacts? (e.g. removing the word `no', which is a strong indicator for \emph{contradiction} may leave the remaining dataset with this word mostly appearing in the \emph{neutral} or \emph{entailment} classes thus create new artifact) and 2) Will DS methods prevent the models to learn specific inference phenomena (e.g. `animal' is a hypernym of `dog')?
First of all, different from \cite{gururangan2018annotation} which only considered unigram patterns, our artificial patterns are mostly multi-word patterns rather than unigram patterns
as the former usually has larger concurrent probability $\mathrm{p}(l|b)$ as shown in Table \ref{showpatterns}. 
Our intention is to use the multi-word patterns to capture the specific ways of expression (human artifacts), rather than single words, of the human annotators. 
For the first concern, instead of filtering the unigram `no', we prefer removing multi-word patterns which contain `no', such as `There are no' or `no \# on' for MultiNLI as shown in Table \ref{showpatterns}. 
For the hypernym mentioned in the second concern, as we prefer filtering multi-word patterns like `The dogs are \# on', we would not deliberately filter the unigram `dog' unless adopting very aggressive DS strategy ($\lambda=0.5$) in both SNLI and MultiNLI. 

%\subsection{Instance Reweighting}
%Down-sampling methods inevitably downgrade the model inference ability as the shrink of training sets. So we intended to figure out a instance reweighting method without sacrificing the training dataset size. 
%We adopt the reweighting setting from \cite{clark2019don}. First we train a hypothesis-only classifier using RoBerta model whose output $b$ is a normalized scores for the three-way classification. $b_i{y_i}$ is the possibility of the hypothesis-only model assigns to the correct label $y_i$ for $i$-th training example. Then we trained the models in a weighted version of data, where the weight for the $i$-th training example is (1-$b_i{y_i}$). In this way, we encourage the models to focus on the examples where the hypothesis-only models get wrong.

\subsection{Adversarial Debiasing}
Since the hypothesis-only bias comes solely from the hypothesis sentence, we wonder if it is possible to get rid of these biases via debiasing the hypothesis sentence vector.
More specifically, we focus on the `sentence vector-based models' \footnote{It would be more challenging to manipulate the gradients in the non-sentence vector-based models, e.g. models which contain interactions between hypothesis and premise sentence encoders like \cite{chen2017neural}. We leave this to the future work.} category as defined on SNLI's web page\footnote{https://nlp.stanford.edu/projects/snli/}. Notably the idea of debiasing NLI via adversarial training has been proposed before \cite{belinkov2019don,belinkov2018mitigating}. We hereby briefly introduce how we implement our adversarial training and how we incorporate instance reweighting method in this framework.

In the following experiments, we use the full training sets without any down-sampling. We use the InferSent \cite{conneau2017supervised} (biLSTM with max pooling) model as the benchmark sentence encoder.

\begin{figure}[t]
\begin{center}
\includegraphics[width=0.8\linewidth]{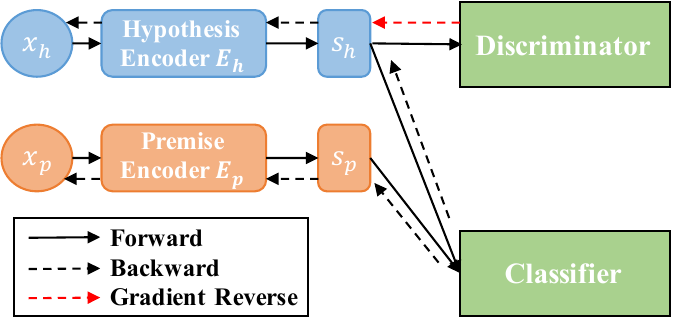}
\end{center}
\caption{The illustration of the sentence-level debiasing framework, which is elaborated in Sec \ref{sec:adver_intro}. }\label{adversarial}
\end{figure}

\subsubsection{Adversarial Debiasing Framework} \label{sec:adver_intro}
As shown in Fig \ref{adversarial}, given the outputs $s_h=\mathrm{E}_h(x_h), s_p=\mathrm{E}_s(x_s)$ of hypothesis and premise encoders $\mathrm{E}_h, \mathrm{E}_p$, we are interested in predicting the NLI label $y$ using a classifier $\mathrm{C}$, $\mathrm{p}_{\mathrm{C}}(y|s_h=\mathrm{E}_h(x_h), s_p=\mathrm{E}_s(x_s))$. 
In addition, 
%we try to prevent the model `benefit' from the hypothesis-only bias by fitting the bias patterns in the hypotheses. So 
we train a hypothesis-only discriminator trying to predict the correct label $y$ solely from the hypothesis sentence representation $s_h$ by modeling $\mathrm{p}_{\mathrm{D}}(y|s_h=\mathrm{E}_h(x_h))$. 
%we try to prevent the model `benefit' from the hypothesis-only bias by fitting the bias patterns in the hypotheses
We formulate the training process in the adversarial setting by a min-max game. Specifically we train the discriminator $\mathrm{D}$ to predict the label using only hypothesis sentence vector. Additionally we train the sentence encoder $\mathrm{E}_h$, $\mathrm{E}_p$ and the classifier $\mathrm{C}$ to fool the discriminator $\mathrm{D}$ without hurting inference ability. $\gamma$ is a hyper-parameter which controls the degree of debiasing.
\begin{equation}
\begin{aligned}
    \min_{\mathrm{E}_h,\mathrm{E}_p,\mathrm{C}} \max_{\mathrm{D}} \mathbbm{E}_{x_h,x_p,y \sim \mathrm{p}(X,Y)}[\gamma \log \mathrm{p}_\mathrm{D}(y|\mathrm{E}_h(x_h)) - 
    \\ \log \mathrm{p}_\mathrm{C}(y|\mathrm{E}_h(x_h), \mathrm{E}_p(x_p))]
    \label{eq:minmax}
\end{aligned}
\end{equation}
 
%In our experiments, we set $\gamma=3$ and $\gamma=1$ for models trained on SNLI and MultiNLI respectively.

%\subsection{Training Process}
We train the encoders, discriminator and classifier in Eq \ref{eq:minmax} together with a gradient reversal layer \cite{ganin2016domain} as shown in Fig \ref{adversarial}.
We negate the gradients from the discriminator $\mathrm{D}$ (red arrow in Fig \ref{adversarial}) to push the hypothesis encoder $\mathrm{E}_h$ to the opposite direction while update its parameters.
The usage of gradient reversal layer makes it easier to optimize the min-max game in Eq \ref{eq:minmax} \cite{xie2017controllable,chen2018adversarial} than training the two adversarial components alternately like Generative Adversarial Nets (GANs) \cite{goodfellow2014generative}.
We update the model parameters $\theta$ by gradient descending ($m$ is the batch size):
\begin{equation}
\small
    \theta_{\mathrm{D}}^{new} = \theta_{\mathrm{D}}^{old} - \frac{1}{m} \sum_{i=1}^m \nabla_{\theta_{\mathrm{D}}} [\log \mathrm{p}_{\mathrm{D}}(y^i|\mathrm{E}_h(x_h^i))]
\label{eq:d}
\end{equation}
\begin{equation}
\small
    \theta_{\mathrm{C}}^{new} = \theta_{\mathrm{C}}^{old} - \frac{1}{m} \sum_{i=1}^m \nabla_{\theta_{\mathrm{C}}}[\log \mathrm{p}_{\mathrm{C}}(y^i|\mathrm{E}_h(x_h^i),\mathrm{E}_p(x_p^i))]
\label{eq:c}
\end{equation}
%\begin{equation}
%\small
%   \theta_{\mathrm{E}_p}^{new} = \theta_{\mathrm{E}_p}^{old} - \frac{1}{m} \sum_{i=1}^m 
%    \nabla_{\theta_{\mathrm{E}_p}}[\log % \mathrm{p}_{\mathrm{C}}(y^i|\mathrm{E}_h(x_h^i),\mathrm{E}_p(x_p^i))]
%\end{equation}
\begin{equation}
\small
\begin{aligned}
   \theta_{\mathrm{E}_h}^{new} = \theta_{\mathrm{E}_h}^{old} - \frac{1}{m} \sum_{i=1}^m 
    \nabla_{\theta_{\mathrm{E}_h}}[\log \mathrm{p}_{\mathrm{C}}(y^i|\mathrm{E}_h(x_h^i),\mathrm{E}_p(x_p^i))]
    \\
 \underbrace{+ \frac{\gamma}{m} \sum_{i=1}^m 
    \nabla_{\theta_{\mathrm{E}_h}}[\log \mathrm{p}_{\mathrm{D}}(y^i|\mathrm{E}_h(x_h^i))]
    }_{\textbf{gradient reverse}}
\end{aligned}
\label{eq:theta_eh}
\end{equation}

\subsubsection{Guidance from Artificial Patterns}\label{sec:guidance}
%As shown in .., the baseline debiasing model does not work well because 
%As shown in Table \ref{showpatterns}, the training sets are not distributed evenly in terms of bias patterns, some instances are more biased than others, which is exactly the reason why the down-sampling methods in the `debias' mode (Table \ref{tb:train_on_snli} and \ref{tb:train_on_mnli}) are helpful to deal with the hypothesis-only bias. 
The artificial patterns turns out to be useful guidances for both the discriminator $\mathrm{D}$ and the classifier $\mathrm{C}$ as they indicate whether an instance is biased or not.
%, which are the keys to the success of the proposed adversarial debiasing method. 
We thus reweight the training instances in the training set based on the division of `biased' and `unbiased' training subsets. 

\noindent \textbf{Guidance for Discriminator}:
During the adversarial process, we optimize the discriminator $\mathrm{D}$ by maximizing the log likelihood loss like Eq \ref{eq:d}.
We find increasing the weights of the biased instances in the training set is of great help to the adversarial debiasing model. Because in this way, the discriminator can learn more from the biased instances to better fit the hypothesis-only bias. The whole adversarial debiasing training process could benefit from a stronger hypothesis-only discriminator.
Formally, we replace negative log likelihood loss function in Eq \ref{eq:d} with a weighted loss function: 
\begin{equation}
\small
\begin{aligned}
    \frac{1}{m} \sum_{i=1}^{m} [\mathbbm{1}\{(x^i,y^i) \in \mathcal{D}_{unbias}\} \log  \mathrm{p}_{\mathrm{D}}(y^i|\mathrm{E}_h(x_h^i)) + \\
    \alpha_1 * \mathbbm{1}\{(x^i,y^i) \in \mathcal{D}_{bias}\} \log  \mathrm{p}_{\mathrm{D}}(y^i|\mathrm{E}_h(x_h^i))]
\end{aligned}
\label{eq:d-reweight}
\end{equation}
%The principle is to increase the weight of the biased instances while training the discriminator $D$ so that it can fit the biased instance better
, where $\mathcal{D} = \mathcal{D}_{unbias} \cup \mathcal{D}_{bias}$ denotes the whole training corpus. The division of biased and unbiased training subsets depends on the debiasing threshold $\lambda$ (just like the down-sampling threhold in Table \ref{tb:train_on_snli} and \ref{tb:train_on_mnli}). $\alpha_1 \geq 1$ is a hyper-parameter which reflects the attention on biased instances for the hypothesis-only discriminator.

\noindent \textbf{Guidance for Classifier}: Similar to the re-weighting method in Eq \ref{eq:d-reweight}, we also apply the re-weighting strategy on the parameter learning for the inference classifier in Eq \ref{eq:c}. We hope the classifier can capture the concrete semantics in NLI instead of over-fitting the artificial patterns in the hypotheses. Thus we increase the weights of the unbiased training subset in the loss function of Eq \ref{eq:c}.
\begin{equation}
\small
\begin{aligned}
    \frac{1}{m} \sum_{i=1}^{m} [\mathbbm{1}\{(x^i,y^i) \in \mathcal{D}_{bias}\} \log  \mathrm{p}_{\mathrm{C}}(y^i|\mathrm{E}_h(x_h^i),\mathrm{E}_h(x_p^i)) + \\
    \alpha_2 * \mathbbm{1}\{(x^i,y^i) \in \mathcal{D}_{unbias}\} \log  \mathrm{p}_{\mathrm{C}}(y^i|\mathrm{E}_h(x_h^i),\mathrm{E}_h(x_p^i))]
\end{aligned}
\label{eq:c-reweight}
\end{equation}
, where $\alpha_2 \geq 1$ is a threshold to control the attention the models pay on the unbiased instances.

%For the results in Table \ref{tb:train_on_adv}, we set $\alpha_1=5,\alpha_2=5$, besides we use $\lambda=0.7$ as the threshold for splitting $\mathcal{D}_{bias}$ and $\mathcal{D}_{unbias}$ in Eq \ref{eq:d-reweight} and \ref{eq:c-reweight}. 

\subsubsection{Training Details}\label{sec:traindetails}
Apart from the weighted loss functions guided by the artificial patterns, we also investigate the following two techniques in the adversarial training process.

\noindent \textbf{Multiple discriminators}: The min-max game in Eq \ref{eq:minmax} could benefit from stronger discriminators. So we try $k \in \{1,2,3\}$ discriminators to enhance its ability to do hypothesis-only classifications. In our experiments, we find that $k=2$ is the best configuration for the discriminator.

\noindent \textbf{Dynamic reweighting}: For hyper-parameter $\alpha$ ($\alpha_1$ and $\alpha_2$ in Eq \ref{eq:d-reweight} and Eq \ref{eq:c-reweight} respectively), we find it useful to adjust $\alpha$ dynamically in the training process. $\alpha^{0}$ and $\alpha^t$ represents the initial values we set before training and its value after $t$ training iterations respectively. Additionally we set up a hyper-parameter $\phi$ to control the gap of model accuracies $\delta$ on the easy and hard subsets in the dev set. 
\begin{equation}
\alpha^{t+1}=
\left\{
 \begin{array}{lr}
 \max(\alpha^t+\epsilon, \alpha^0), & \delta \geq \phi \\
 \max(\alpha^t-\epsilon, \alpha^0), & \delta < \phi \\
 \end{array}
\right.
\end{equation}
where $\epsilon$ is a hyper-parameter set as 0.5 for models trained on both datasets. Besides, we set $\phi$ as 0.15 and 0.10 for SNLI and MultiNLI respectively.
Notably although we update the hyper-parameters $\alpha_1$ and $\alpha_2$ dynamically in different iterations based on $\phi$, we still select the model which has the best performance on the dev sets as the best model in each run.

\noindent \textbf{Parameter settings}: We use grid search to find the best hyper-parameter settings: $\alpha_1,\alpha_2 \in \{1,3,5,10\}$, $\gamma \in \{0.5,1,3,5\}$ in Eq \ref{eq:d-reweight}, \ref{eq:c-reweight} and Eq \ref{eq:minmax}.
We also try $\lambda \in \{0.5,0.6,0.7,0.8\}$ as the threshold to split $\mathcal{D}_{bias}$ and $\mathcal{D}_{unbias}$ in Eq \ref{eq:d-reweight} and \ref{eq:c-reweight}. 
Specifically, we treat the instances which contain the artificial patterns in $\mathrm{H}$(3,3,$\lambda$) (Sec \ref{extraction}, Footnote \ref{fn:mandt}) as $\mathcal{D}_{bias}$, and the remaining instances as $\mathcal{D}_{unbias}$.
For the results in Table \ref{tb:train_on_adv}, we set $\gamma=3$ and $\gamma=1$ for SNLI and MultiNLI respectively. For both datasets, we set $\alpha_1=5,\alpha_2=5$ as well as $\lambda=0.7$ as the threshold for separating the biased and unbiased subsets in Eq \ref{eq:d-reweight} and \ref{eq:c-reweight}. 
For a fair comparison, we do not tune any hyper-parameter in the InferSent encoder, the learning rate and the optimizer setting. The results of `dInferSent' and its variations in Table \ref{tb:train_on_adv} comes from 5 independent runs with random initialization.

\begin{table}[]
\small
\begin{center}
\begin{tabular}{lcccc}
\hline 
\textbf{Model}  & \textbf{Full} & \textbf{Easy} & \textbf{Hard} & $\Delta^{Hard}_{Easy}(\downarrow)$ \\ \hline
InferSent & \underline{\textbf{84.5}} & \underline{\textbf{97.2}} & 58.9 & 38.3 \\
InferSent+DS($\lambda$=0.8) & 82.9 & 90.4 & 60.0 & 30.4 \\ 
InferSent+Guidance & 84.1 & 95.5 & \textbf{61.7} & 33.8 \\ \hline
dInferSent & 81.6 & \textbf{92.5} & 59.9 & 32.6 \\
+Guidance & \textbf{82.2} & 86.9 & 63.3 & 23.6 \\
+Guidance+Reweight & 80.9 & 78.2 & \underline{\textbf{67.3}} & 10.9 \\
\hline
\end{tabular}
\\(a) InferSent trained on SNLI

\begin{tabular}{lcccc}
\hline 
\textbf{Model}  & \textbf{Full} & \textbf{Easy} & \textbf{Hard} & $\Delta^{Hard}_{Easy}(\downarrow)$ \\ \hline
InferSent & \underline{\textbf{70.4}} & \underline{\textbf{92.7}} & 54.4 & 38.3 \\
InferSent+DS($\lambda$=0.8) & 69.9 & 91.4 & 53.6 & 37.8 \\
InferSent+Guidance & 70.1 & 92.1 & \textbf{54.9} & 37.2 \\ \hline
dInferSent & \textbf{68.8} & \textbf{91.1} & 54.7 & 36.4 \\
+Guidance & 68.0 & 87.9 & 55.3 & 32.6 \\
+Guidance+Reweight & 66.5 & 79.4 & \underline{\textbf{58.8}} & 20.6 \\
\hline
\end{tabular}
\\(b) InferSent trained on MultiNLI

\end{center}
%}
\caption{
The comparison of InferSent (baseline), InferSent+DS (downsampling) and dInferSent (adversarial debiasing) on SNLI test set and MultiNLI mismatched dev set respectively. We choose the down-sampling (DS) method with $\lambda=0.8$ because it performs best on the hard subsets. The `Guidance' and `Reweight' methods are elaborated in Sec \ref{sec:guidance} and Sec \ref{sec:traindetails} respectively.
}\label{tb:train_on_adv}
\end{table}

\subsubsection{Discussions}
From Table \ref{tb:train_on_adv},
we observe that although the performance gap between the easy and hard subsets is reduced to some extent by the vanilla dInferSent models in both SNLI and MultiNLI. The model still does not reach our expectation to lower the gap between hard and easy sets. We assume this is because the denoising discriminator in Fig \ref{adversarial} somewhat impedes the inference ability of the NLI models as it may disturb the hypothesis sentence encoder especially when the sentences do not contain hypothesis-only bias. 
The explicit guidance (`+Guidance') from the artificial patterns alleviates this issue in both datasets as in this way the discriminator pays more attention on the potentially biased instances thus has smaller influence on the hard instances in the training procedure. These models achieve higher accuracies on the hard  subset than the baseline models in both datasets.
The `reweight' trick in Sec \ref{sec:traindetails} greatly reduces the performance gap between the easy and hard sets as it dynamically adjusts the debiasing strategies (i.e. the weight of training instances in Eq \ref{eq:d-reweight} and \ref{eq:c-reweight}). We see similar reweighting tricks also work in the debiasing of relation extraction \cite{liu-etal-2017-soft}.
%The debiasing guided by the bias patterns is another feasible way to mitigate the bias without downsampling the training sets.
% \nocite{*}

\section{Related Work}
The bias in the data annotation exists in many tasks, e.g. lexical inference \cite{levy2015supervised}, visual question answering \cite{goyal2017making}, ROC story cloze \cite{DBLP:conf/acl/CaiTG17} etc.
The NLI models are shown to be sensitive to the compositional features in premises and hypotheses~\cite{nie2018analyzing}, data permutations \cite{schluter2018data,wang2018if} and vulnerable to adversarial examples \cite{iyyer2018adversarial,minervini2018adversarially,GlocknerSG18,liu-etal-2020-empirical} and crafted stress test \cite{geiger2018stress,DBLP:conf/coling/NaikRSRN18}.  
\cite{RudingerMD17} showed hypothesis in SNLI has the evidence of gender, racial and religious stereotypes, etc. 
\cite{sanchez2018behavior} analysed the behaviour of NLI models and the factors to be more robust.
\cite{feng2019misleading} discussed how to use partial-input baseline (hypothesis-only classifier in NLI) in future dataset creation.
\cite{clark2019don} uses an ensemble-based method to mitigate known bias.
The InferSent model, which served as an important baseline in this paper, are found to achieve superb performance on SNLI by word-level heuristics \cite{dasgupta2018evaluating}.
\cite{ding-etal-2020-discriminatively} proposed shows the advantages of generative training in increasing NLI robustness.

\cite{maccartney2009natural} first revealed the difficulties of natural language inference model with bag-of-words models.
Different from the artificial patterns we used in this paper, other artifact evidence includes sentence occurrence \cite{zhang2019selection}, syntactic heuristics between hypotheses and premises \cite{mccoy2019right} and black-box clues derived from neural models \cite{gururangan2018annotation,poliak2018hypothesis,he2019unlearn,karimi-mahabadi-etal-2020-end}.

The adversarial debiasing training proposed in this paper is inspired by the success of Generative Adversarial Networks (GANs) \cite{goodfellow2014generative}.
Several works on learning encoders which are invariant to certain properties of text and image \cite{chen2018adversarial,zhang2017aspect,xie2017controllable,moyer2018evading,jaiswal2018unsupervised} in the adversarial settings. 
%Another line of researches aims at removing sensitive attributes from intermediate representations in order to achieve fair classification by adversarial training \cite{edwards2015censoring,louizos2015variational,zhang2018mitigating,elazar2018adversarial}.
%They listed some biased unigrams like `sleeping', `outside' in the hypotheses.
%However, we find that multi-word bias patterns also contribute to the hypothesis-only bias, which is helpful to avoid such bias when annotating new datasets. Although \citet{gururangan2018annotation} had little faith in down-sampling methods, our work shows that we can mitigate, as least to some extend, the hypothesis-only bias by adopting appropriate down-sampling strategy.  

\section{Conclusion}
In this study, we show that the hypothesis-only bias in trained NLI models mainly comes from unevenly distributed surface patterns, which could be used to identify hard and easy instances for more convincing re-evaluation on currently overestimated NLI models.
The attempts to mitigate the bias are meaningful as such bias not only makes NLI models fragile to adversarial examples. 
We try to mitigate this bias by removing those artificial patterns in the training sets, with experiments showing that it is a feasible way to alleviate the bias under proper down-sampling methods.
We also show that adversarial debiasing with the guidance from the harvested artificial patterns is a feasible way to mitigate the hypothesis-only bias for sentence vector-based NLI models.

\section*{Acknowledgments}
We would like to thank the anonymous reviewers for their valuable suggestions.  
This work is supported by the National Science Foundation of China under Grant No. 61751201, No. 61772040, No. 61876004.
The corresponding authors of this paper are Baobao Chang and Zhifang Sui.

\section{Bibliographical References}
\label{main:ref}

\bibliographystyle{lrec}
\bibliography{lrec2020}

\begin{thebibliography}{}

\bibitem[\protect\citename{Bowman \bgroup et al.\egroup }2015]{snli:emnlp2015}
Bowman, S.~R., Angeli, G., Potts, C., and Manning, C.~D.
\newblock (2015).
\newblock A large annotated corpus for learning natural language inference.
\newblock In {\em EMNLP}. Association for Computational Linguistics.

\bibitem[\protect\citename{Williams \bgroup et al.\egroup }2018]{mnli:N18-1101}
Williams, A., Nangia, N., and Bowman, S.
\newblock (2018).
\newblock A broad-coverage challenge corpus for sentence understanding through
  inference.
\newblock In {\em NAACL}, pages 1112--1122. Association for Computational
  Linguistics.

\end{thebibliography}


\begin{thebibliography}{}

\bibitem[\protect\citename{Belinkov \bgroup et al.\egroup
  }2018]{belinkov2018mitigating}
Belinkov, Y., Poliak, A., Shieber, S.~M., and Van~Durme, B.
\newblock (2018).
\newblock Mitigating bias in natural language inference using adversarial
  learning.

\bibitem[\protect\citename{Belinkov \bgroup et al.\egroup
  }2019]{belinkov2019don}
Belinkov, Y., Poliak, A., Shieber, S.~M., Van~Durme, B., and Rush, A.~M.
\newblock (2019).
\newblock Don't take the premise for granted: Mitigating artifacts in natural
  language inference.
\newblock {\em arXiv preprint arXiv:1907.04380}.

\bibitem[\protect\citename{Cai \bgroup et al.\egroup
  }2017]{DBLP:conf/acl/CaiTG17}
Cai, Z., Tu, L., and Gimpel, K.
\newblock (2017).
\newblock Pay attention to the ending: Strong neural baselines for the {ROC}
  story cloze task.
\newblock In {\em ACL}.

\bibitem[\protect\citename{Chen \bgroup et al.\egroup }2017a]{chen2017neural}
Chen, Q., Zhu, X., Ling, Z.-H., Inkpen, D., and Wei, S.
\newblock (2017a).
\newblock Neural natural language inference models enhanced with external
  knowledge.
\newblock {\em arXiv preprint arXiv:1711.04289}.

\bibitem[\protect\citename{Chen \bgroup et al.\egroup }2017b]{chen2016enhanced}
Chen, Q., Zhu, X., Ling, Z., Wei, S., Jiang, H., and Inkpen, D.
\newblock (2017b).
\newblock Enhanced {LSTM} for natural language inference.
\newblock In {\em ACL}.

\bibitem[\protect\citename{Chen \bgroup et al.\egroup
  }2018]{chen2018adversarial}
Chen, X., Sun, Y., Athiwaratkun, B., Cardie, C., and Weinberger, K.
\newblock (2018).
\newblock Adversarial deep averaging networks for cross-lingual sentiment
  classification.
\newblock {\em TACL}, 6:557--570.

\bibitem[\protect\citename{Clark \bgroup et al.\egroup }2019]{clark2019don}
Clark, C., Yatskar, M., and Zettlemoyer, L.
\newblock (2019).
\newblock Don't take the easy way out: Ensemble based methods for avoiding
  known dataset biases.
\newblock {\em arXiv preprint arXiv:1909.03683}.

\bibitem[\protect\citename{Conneau \bgroup et al.\egroup
  }2017]{conneau2017supervised}
Conneau, A., Kiela, D., Schwenk, H., Barrault, L., and Bordes, A.
\newblock (2017).
\newblock Supervised learning of universal sentence representations from
  natural language inference data.
\newblock In {\em EMNLP}, pages 670--680.

\bibitem[\protect\citename{Dagan \bgroup et al.\egroup }2006]{dagan2006pascal}
Dagan, I., Glickman, O., and Magnini, B.
\newblock (2006).
\newblock The pascal recognising textual entailment challenge.
\newblock pages 177--190. Springer.

\bibitem[\protect\citename{Dagan \bgroup et al.\egroup
  }2013]{DBLP:series/synthesis/2013Dagan}
Dagan, I., Roth, D., Sammons, M., and Zanzotto, F.~M.
\newblock (2013).
\newblock {\em Recognizing Textual Entailment: Models and Applications}.
\newblock Synthesis Lectures on Human Language Technologies.

\bibitem[\protect\citename{Dasgupta \bgroup et al.\egroup
  }2018]{dasgupta2018evaluating}
Dasgupta, I., Guo, D., Stuhlm{\"u}ller, A., Gershman, S.~J., and Goodman, N.~D.
\newblock (2018).
\newblock Evaluating compositionality in sentence embeddings.
\newblock {\em arXiv preprint arXiv:1802.04302}.

\bibitem[\protect\citename{Devlin \bgroup et al.\egroup }2018]{devlin2018bert}
Devlin, J., Chang, M.-W., Lee, K., and Toutanova, K.
\newblock (2018).
\newblock Bert: Pre-training of deep bidirectional transformers for language
  understanding.
\newblock {\em arXiv preprint arXiv:1810.04805}.

\bibitem[\protect\citename{Ding \bgroup et al.\egroup
  }2020]{ding-etal-2020-discriminatively}
Ding, X., Liu, T., Chang, B., Sui, Z., and Gimpel, K.
\newblock (2020).
\newblock Discriminatively-{T}uned {G}enerative {C}lassifiers for {R}obust
  {N}atural {L}anguage {I}nference.
\newblock In {\em Proceedings of the 2020 Conference on Empirical Methods in
  Natural Language Processing (EMNLP)}, pages 8189--8202, Online, November.
  Association for Computational Linguistics.

\bibitem[\protect\citename{Feng \bgroup et al.\egroup
  }2019]{feng2019misleading}
Feng, S., Wallace, E., and Boyd-Graber, J.
\newblock (2019).
\newblock Misleading failures of partial-input baselines.
\newblock {\em arXiv preprint arXiv:1905.05778}.

\bibitem[\protect\citename{Ganin \bgroup et al.\egroup }2016]{ganin2016domain}
Ganin, Y., Ustinova, E., Ajakan, H., Germain, P., Larochelle, H., Laviolette,
  F., Marchand, M., and Lempitsky, V.
\newblock (2016).
\newblock Domain-adversarial training of neural networks.
\newblock {\em JMLR}, 17(1):2096--2030.

\bibitem[\protect\citename{Geiger \bgroup et al.\egroup
  }2018]{geiger2018stress}
Geiger, A., Cases, I., Karttunen, L., and Potts, C.
\newblock (2018).
\newblock Stress-testing neural models of natural language inference with
  multiply-quantified sentences.
\newblock {\em arXiv preprint arXiv:1810.13033}.

\bibitem[\protect\citename{Glockner \bgroup et al.\egroup }2018]{GlocknerSG18}
Glockner, M., Shwartz, V., and Goldberg, Y.
\newblock (2018).
\newblock Breaking {NLI} systems with sentences that require simple lexical
  inferences.
\newblock In {\em ACL}, pages 650--655.

\bibitem[\protect\citename{Goodfellow \bgroup et al.\egroup
  }2014]{goodfellow2014generative}
Goodfellow, I., Pouget-Abadie, J., Mirza, M., Xu, B., Warde-Farley, D., Ozair,
  S., Courville, A., and Bengio, Y.
\newblock (2014).
\newblock Generative adversarial nets.
\newblock In {\em NIPS}, pages 2672--2680.

\bibitem[\protect\citename{Goyal \bgroup et al.\egroup }2017]{goyal2017making}
Goyal, Y., Khot, T., Summers-Stay, D., Batra, D., and Parikh, D.
\newblock (2017).
\newblock Making the {V} in {VQA} matter: Elevating the role of image
  understanding in visual question answering.
\newblock In {\em CVPR}.

\bibitem[\protect\citename{Gururangan \bgroup et al.\egroup
  }2018]{gururangan2018annotation}
Gururangan, S., Swayamdipta, S., Levy, O., Schwartz, R., Bowman, S.~R., and
  Smith, N.~A.
\newblock (2018).
\newblock Annotation artifacts in natural language inference data.
\newblock In {\em NAACL}, pages 107--112.

\bibitem[\protect\citename{He \bgroup et al.\egroup }2019]{he2019unlearn}
He, H., Zha, S., and Wang, H.
\newblock (2019).
\newblock Unlearn dataset bias in natural language inference by fitting the
  residual.
\newblock {\em arXiv preprint arXiv:1908.10763}.

\bibitem[\protect\citename{Iyyer \bgroup et al.\egroup
  }2018]{iyyer2018adversarial}
Iyyer, M., Wieting, J., Gimpel, K., and Zettlemoyer, L.
\newblock (2018).
\newblock Adversarial example generation with syntactically controlled
  paraphrase networks.
\newblock In {\em NAACL}, pages 1875--1885.

\bibitem[\protect\citename{Jaiswal \bgroup et al.\egroup
  }2018]{jaiswal2018unsupervised}
Jaiswal, A., Wu, R.~Y., Abd-Almageed, W., and Natarajan, P.
\newblock (2018).
\newblock Unsupervised adversarial invariance.
\newblock In {\em NIPS}, pages 5097--5107.

\bibitem[\protect\citename{Joulin \bgroup et al.\egroup
  }2016]{Joulin2016FastText}
Joulin, A., Grave, E., Bojanowski, P., Douze, M., J{\'e}gou, H., and Mikolov,
  T.
\newblock (2016).
\newblock Fasttext. zip: Compressing text classification models.
\newblock {\em arXiv preprint arXiv:1612.03651}.

\bibitem[\protect\citename{Karimi~Mahabadi \bgroup et al.\egroup
  }2020]{karimi-mahabadi-etal-2020-end}
Karimi~Mahabadi, R., Belinkov, Y., and Henderson, J.
\newblock (2020).
\newblock End-to-end bias mitigation by modelling biases in corpora.
\newblock In {\em Proceedings of the 58th Annual Meeting of the Association for
  Computational Linguistics}, pages 8706--8716, Online, July. Association for
  Computational Linguistics.

\bibitem[\protect\citename{Levy \bgroup et al.\egroup
  }2015]{levy2015supervised}
Levy, O., Remus, S., Biemann, C., and Dagan, I.
\newblock (2015).
\newblock Do supervised distributional methods really learn lexical inference
  relations?
\newblock In {\em NAACL}, pages 970--976.

\bibitem[\protect\citename{Lin \bgroup et al.\egroup }2017]{lin2017structured}
Lin, Z., Feng, M., Santos, C. N.~d., Yu, M., Xiang, B., Zhou, B., and Bengio,
  Y.
\newblock (2017).
\newblock A structured self-attentive sentence embedding.
\newblock {\em arXiv preprint arXiv:1703.03130}.

\bibitem[\protect\citename{Liu \bgroup et al.\egroup }2017]{liu-etal-2017-soft}
Liu, T., Wang, K., Chang, B., and Sui, Z.
\newblock (2017).
\newblock A soft-label method for noise-tolerant distantly supervised relation
  extraction.
\newblock In {\em Proceedings of the 2017 Conference on Empirical Methods in
  Natural Language Processing}, pages 1790--1795, Copenhagen, Denmark,
  September. Association for Computational Linguistics.

\bibitem[\protect\citename{Liu \bgroup et al.\egroup }2019]{liu2019roberta}
Liu, Y., Ott, M., Goyal, N., Du, J., Joshi, M., Chen, D., Levy, O., Lewis, M.,
  Zettlemoyer, L., and Stoyanov, V.
\newblock (2019).
\newblock Roberta: A robustly optimized bert pretraining approach.
\newblock {\em arXiv preprint arXiv:1907.11692}.

\bibitem[\protect\citename{Liu \bgroup et al.\egroup
  }2020]{liu-etal-2020-empirical}
Liu, T., Xin, Z., Ding, X., Chang, B., and Sui, Z.
\newblock (2020).
\newblock An empirical study on model-agnostic debiasing strategies for robust
  natural language inference.
\newblock In {\em Proceedings of the 24th Conference on Computational Natural
  Language Learning}, pages 596--608, Online, November. Association for
  Computational Linguistics.

\bibitem[\protect\citename{Luo \bgroup et al.\egroup }2018]{luo2018leveraging}
Luo, F., Liu, T., He, Z., Xia, Q., Sui, Z., and Chang, B.
\newblock (2018).
\newblock Leveraging gloss knowledge in neural word sense disambiguation by
  hierarchical co-attention.
\newblock In {\em Proceedings of the 2018 Conference on Empirical Methods in
  Natural Language Processing}, pages 1402--1411.

\bibitem[\protect\citename{MacCartney and Manning}2009]{maccartney2009natural}
MacCartney, B. and Manning, C.~D.
\newblock (2009).
\newblock {\em Natural language inference}.
\newblock Citeseer.

\bibitem[\protect\citename{McCoy \bgroup et al.\egroup }2019]{mccoy2019right}
McCoy, R.~T., Pavlick, E., and Linzen, T.
\newblock (2019).
\newblock Right for the wrong reasons: Diagnosing syntactic heuristics in
  natural language inference.
\newblock {\em arXiv preprint arXiv:1902.01007}.

\bibitem[\protect\citename{Minervini and
  Riedel}2018]{minervini2018adversarially}
Minervini, P. and Riedel, S.
\newblock (2018).
\newblock Adversarially regularising neural {NLI} models to integrate logical
  background knowledge.
\newblock In {\em CoNLL}, pages 65--74.

\bibitem[\protect\citename{Moyer \bgroup et al.\egroup }2018]{moyer2018evading}
Moyer, D., Gao, S., Brekelmans, R., Steeg, G.~V., and Galstyan, A.
\newblock (2018).
\newblock Evading the adversary in invariant representation.
\newblock {\em arXiv preprint arXiv:1805.09458}.

\bibitem[\protect\citename{Naik \bgroup et al.\egroup
  }2018]{DBLP:conf/coling/NaikRSRN18}
Naik, A., Ravichander, A., Sadeh, N., Ros{\'{e}}, C.~P., and Neubig, G.
\newblock (2018).
\newblock Stress test evaluation for natural language inference.
\newblock In {\em COLING}, pages 2340--2353.

\bibitem[\protect\citename{Nie \bgroup et al.\egroup }2019a]{nie2018analyzing}
Nie, Y., Wang, Y., and Bansal, M.
\newblock (2019a).
\newblock Analyzing compositionality-sensitivity of {NLI} models.
\newblock {\em AAAI}.

\bibitem[\protect\citename{Nie \bgroup et al.\egroup
  }2019b]{nie2019adversarial}
Nie, Y., Williams, A., Dinan, E., Bansal, M., Weston, J., and Kiela, D.
\newblock (2019b).
\newblock Adversarial nli: A new benchmark for natural language understanding.
\newblock {\em arXiv preprint arXiv:1910.14599}.

\bibitem[\protect\citename{Parikh \bgroup et al.\egroup
  }2016]{parikh2016decomposable}
Parikh, A.~P., T{\"{a}}ckstr{\"{o}}m, O., Das, D., and Uszkoreit, J.
\newblock (2016).
\newblock A decomposable attention model for natural language inference.
\newblock In {\em EMNLP}, pages 2249--2255.

\bibitem[\protect\citename{Poliak \bgroup et al.\egroup
  }2018]{poliak2018hypothesis}
Poliak, A., Naradowsky, J., Haldar, A., Rudinger, R., and Durme, B.~V.
\newblock (2018).
\newblock Hypothesis only baselines in natural language inference.
\newblock In {\em *SEM@NAACL-HLT}, pages 180--191.

\bibitem[\protect\citename{Rudinger \bgroup et al.\egroup }2017]{RudingerMD17}
Rudinger, R., May, C., and Durme, B.~V.
\newblock (2017).
\newblock Social bias in elicited natural language inferences.
\newblock In {\em EthNLP@EACL}, pages 74--79.

\bibitem[\protect\citename{Sanchez \bgroup et al.\egroup
  }2018]{sanchez2018behavior}
Sanchez, I., Mitchell, J., and Riedel, S.
\newblock (2018).
\newblock Behavior analysis of nli models: Uncovering the influence of three
  factors on robustness.
\newblock In {\em EMNLP}, volume~1, pages 1975--1985.

\bibitem[\protect\citename{Schluter and Varab}2018]{schluter2018data}
Schluter, N. and Varab, D.
\newblock (2018).
\newblock When data permutations are pathological: the case of neural natural
  language inference.
\newblock In {\em EMNLP}, pages 4935--4939.

\bibitem[\protect\citename{Tsuchiya}2018]{tsuchiya2018performance}
Tsuchiya, M.
\newblock (2018).
\newblock Performance impact caused by hidden bias of training data for
  recognizing textual entailment.
\newblock In {\em LREC}.

\bibitem[\protect\citename{Wang \bgroup et al.\egroup }2018]{wang2018if}
Wang, H., Sun, D., and Xing, E.~P.
\newblock (2018).
\newblock What if we simply swap the two text fragments? a straightforward yet
  effective way to test the robustness of methods to confounding signals in
  nature language inference tasks.
\newblock {\em arXiv preprint arXiv:1809.02719}.

\bibitem[\protect\citename{Wu \bgroup et al.\egroup }2018]{wu2018phrase}
Wu, W., Wang, H., Liu, T., and Ma, S.
\newblock (2018).
\newblock Phrase-level self-attention networks for universal sentence encoding.
\newblock In {\em Proceedings of the 2018 Conference on Empirical Methods in
  Natural Language Processing}, pages 3729--3738.

\bibitem[\protect\citename{Xie \bgroup et al.\egroup
  }2017]{xie2017controllable}
Xie, Q., Dai, Z., Du, Y., Hovy, E., and Neubig, G.
\newblock (2017).
\newblock Controllable invariance through adversarial feature learning.
\newblock In {\em NIPS}.

\bibitem[\protect\citename{Yang \bgroup et al.\egroup
  }2016]{yang2016hierarchical}
Yang, Z., Yang, D., Dyer, C., He, X., Smola, A., and Hovy, E.
\newblock (2016).
\newblock Hierarchical attention networks for document classification.
\newblock In {\em NAACL 2016}, pages 1480--1489.

\bibitem[\protect\citename{Yang \bgroup et al.\egroup }2019]{yang2019xlnet}
Yang, Z., Dai, Z., Yang, Y., Carbonell, J., Salakhutdinov, R., and Le, Q.~V.
\newblock (2019).
\newblock Xlnet: Generalized autoregressive pretraining for language
  understanding.
\newblock {\em arXiv preprint arXiv:1906.08237}.

\bibitem[\protect\citename{Zellers \bgroup et al.\egroup
  }2018]{zellers2018swag}
Zellers, R., Bisk, Y., Schwartz, R., and Choi, Y.
\newblock (2018).
\newblock Swag: A large-scale adversarial dataset for grounded commonsense
  inference.
\newblock {\em arXiv preprint arXiv:1808.05326}.

\bibitem[\protect\citename{Zhang \bgroup et al.\egroup }2017]{zhang2017aspect}
Zhang, Y., Barzilay, R., and Jaakkola, T.
\newblock (2017).
\newblock Aspect-augmented adversarial networks for domain adaptation.
\newblock {\em TACL}, 5:515--528.

\bibitem[\protect\citename{Zhang \bgroup et al.\egroup
  }2019]{zhang2019selection}
Zhang, G., Bai, B., Liang, J., Bai, K., Chang, S., Yu, M., Zhu, C., and Zhao,
  T.
\newblock (2019).
\newblock Selection bias explorations and debias methods for natural language
  sentence matching datasets.
\newblock {\em arXiv preprint arXiv:1905.06221}.

\end{thebibliography}

\section{Language Resource References}
\label{lr:ref}
\bibliographystylelanguageresource{lrec}
\bibliographylanguageresource{languageresource}

\end{document}